\pdfoutput=1
\documentclass[conference, 11pt]{IEEEtran}
\IEEEoverridecommandlockouts
\usepackage[dvipsnames]{xcolor}

\usepackage[square,sort&compress,comma,numbers]{natbib}
\usepackage[utf8]{inputenc} 
\usepackage[T1]{fontenc}    
\usepackage{url}            
\usepackage{booktabs}       
\usepackage{amsfonts}       
\usepackage{nicefrac}       
\usepackage{microtype}      
\usepackage{times}
\usepackage{soul}
\usepackage[utf8]{inputenc}
\usepackage{graphicx}
\usepackage{booktabs}
\usepackage{outlines}
\usepackage{adjustbox}

\usepackage[colorlinks=true,
citecolor=blue,
filecolor=black,
linkcolor=blue,
urlcolor=blue]{hyperref}

\usepackage{microtype}
\usepackage{graphicx}
\usepackage{subfig}
\usepackage{booktabs} 

\usepackage{verbatim}
\usepackage{multirow}
\usepackage{paralist}

\usepackage{booktabs} 
\usepackage{multicol}
\usepackage{diagbox}

\usepackage{here}

\usepackage{amsmath,amssymb,amsfonts,amsbsy,amsfonts,latexsym}
\usepackage{multirow}
\usepackage{makecell}
\usepackage[labelfont=bf,textfont=it,belowskip=0pt,aboveskip=5pt,tableposition=top]{caption}
\usepackage{colortbl}

\definecolor{colorA}{RGB}{189,201,225}
\definecolor{colorB}{RGB}{103,169,207}
\definecolor{colorC}{RGB}{ 28,144,153}
\definecolor{colorD}{RGB}{  1,108, 89}

\newcolumntype{R}{>{\columncolor{gray!40}}r}
\newcolumntype{L}{>{\columncolor{gray!40}}l}
\newcolumntype{C}{>{\columncolor{gray!40}}c}

\usepackage{tabularx,colortbl}
\usepackage{multirow}
\usepackage[normalem]{ulem}
\useunder{\uline}{\ul}{}

\usepackage{enumitem}

\usepackage{xparse}

\DeclareGraphicsExtensions{.pdf,.png}

\usepackage{longtable}
\usepackage{pgfplots}

\usepackage{amsmath}
\usepackage{pifont}

\usepackage{booktabs}       
\usepackage{amsfonts}       
\usepackage{nicefrac}       
\usepackage{microtype}      
\usepackage{xspace}      

\usepackage[utf8]{inputenc} 
\usepackage[T1]{fontenc}    
\usepackage{chngcntr}

\usepackage[utf8]{inputenc}
\usepackage[english]{babel}

\NewDocumentCommand{\var}{O{s} m O{}}{%
  \ensuremath{#1_{#2}^{#3}}
}
\usepackage{siunitx}

\definecolor{light-gray}{gray}{0.80}

\newcommand\eref{Eq.~\ref}
\newcommand\fref{Figure~\ref}
\newcommand\tref{Table~\ref}

\newcommand\ssref{Subsection~\ref}

\definecolor{brickred}{rgb}{0.8, 0.25, 0.33}
\definecolor{brickred2}{rgb}{0.25, 0.8, 0.33}

\begin{document}

\title{Integer-Only Neural Network Quantization Scheme Based on Shift-Batch-Normalization}

\author{
	\large
	Qingyu Guo,
	Yuan Wang,
	Xiaoxin Cui\\
	Peking University\\
	{\tt\small \{guoqinyu, wangyuan, cuixx\}@pku.edu.cn}
}
\maketitle
\pagestyle{plain}

\begin{abstract}
Neural networks are very popular in many areas, but great computing complexity makes it hard to run neural networks on devices with limited resources. To address this problem, quantization methods are used to reduce model size and computation cost, making it possible to use neural networks on embedded platforms or mobile devices. 

In this paper, an integer-only-quantization scheme is introduced. This scheme uses one layer that combines shift-based batch normalization and uniform quantization to implement 4-bit integer-only inference. Without big integer multiplication(which is used in previous integer-only-quantization methods), this scheme can achieve good power and latency efficiency, and is especially suitable to be deployed on co-designed hardware platforms. Tests have proved that this scheme works very well for easy tasks. And for tough tasks, performance loss can be tolerated for its inference efficiency. Our work is available on github\footnote{\href{https://github.com/hguq/IntegerNet}{IntegerNet}}.
\end{abstract}

\section{Introduction}
\label{sec:introduction}

Convolutional Neural Network(CNN) consumes a large amount of computing resources. Even in the inference stage, the computing cost is still not affordable for most mobile and embedded devices. To deploy CNN on these resource-limited devices, there has been a large body of work. Approaches can be roughly categorized as follows.\\

\paragraph{\textbf{Designing more efficient neural network architecture}}
Neural networks are usually recognized to be over-parameterized. Some work focuses on designing more efficient networks that have less parameter and do less calculation. For example, MobileNet\cite{mobile}, SqueezeNet\cite{squeezenet}, ShuffleNet\cite{shufflenet} and so on. Some work tries to prune and compress the whole model. For example, Deep Compression\cite{deepcompression}.

\paragraph{\textbf{Quantization}}
Typically, a neural network calculates with float numbers. According to \cite{energyproblem}, the floating operation requires much more power and latency than integer operations. By approximate float-point operations with integer or fixed-point operations(which is known as "Quantization"), will greatly accelerate the process of inference.
Some work quantizes weight and activations to 8 bit fixed-point, and the performance has no degradation even on tough tasks. Examples are \cite{google,hawqv3}. Other works do extreme quantization, reducing the bit width of activation and weight to 2 and even 1. Examples are TNN\cite{tenary}, BNN\cite{bnn}, Binary Connect\cite{binaryconnect}, XNOR \cite{XNOR}, etc.

\paragraph{\textbf{Architecture-Hardware co-designing}}
By co-designing neural network architecture and hardware architecture, the inference will meet less bottleneck on these specifically designed platforms, bringing great power and latency optimization.
\\

These three methods are not exclusive to each other. For example, architecture-hardware co-designing can be applied to a quantized network, which will harness the advantages of the two methods. Also, quantization can be applied to efficient architecture like MobileNet.

Among the three categories, Quantization is a commonly adopted method to reduce model size as well as improve latency and power performance. But, not all quantization methods are friendly to hardware. The most important problem is that some quantization schemes only quantize part of floating point operations. For hardware platforms that don't support float calculation, these quantized models will not be able to be deployed.

To address this problem, integer-only-quantization schemes are proposed. In these schemes, all floating point operations are converted to integer operations. \cite{google} proposed that, by storing the scale and zero point of each tensor, integer-only quantization scheme can be achieved. \cite{ibert} proposed integer-only-quantization methods for some non-linear operations like softmax and GELU. \cite{hawqv3} proposed a quantization scheme for residual block and batch normalization layer, so ResNet\cite{resnet} architecture can be implemented with only integer operation.

In these integer-only schemes, neural network inference can be carried out with only integer operation, but there is still one thing that can be improved: reducing largely accumulated big activation integer(32bit) back to lower precision integer(8bit or 4bit) requires big integer multiplications. These multiplications may make up a big part of the power consumption of the whole hardware platform.

To address this drawback, we implement an integer-only-quantization scheme without big integer multiplications. Our work is inspired by \cite{bnn}, which leverages the efficiency of shifting operation to do batch normalization.

\section{Quantization Scheme}
\label{sec:quantization scheme}
In this section, our quantization scheme will be explained. Firstly, two problems in uniform quantization method will be discussed. Then we will demonstrate one important empirical fact about batch normalization, which inspired us to solve both problems by using Shift-based Batch Normalization(SBN). Finally, we will introduce a new quantization layer fusing SBN and quantization that only uses integer adding and shifting. The quantization method is called SBNQ, and can be used to quantize signed activations, unsigned activations and weights.
\subsection{Uniform Quantization}
\label{sec:uniform quantization}
In the scheme of integer-only quantization, for one input activition consisted of big integers: $$\mathrm{act_{int}}\in \mathbb{N}$$the target of $\mathrm{N}$-bit quantization is to convert it to integer with limited range\footnote{For signed integer quantization, the most negative number $-2^\mathrm{N-1}$ is not used, the reason is illustrated in \cite{google}}: $$Q(\mathrm{act_{int}})\in [-2^\mathrm{N-1}+1,2^\mathrm{N-1}-1]\cap \mathbb{N}$$

The most used quantization method is uniform quantization, which means the mapping from real domain to integer domain is linear. In uniform quantization, there are several parameters:
\begin{itemize}
	\item The bitwidth of quantization target: $\mathrm{N}$
	\item The lower limit to clip input tensor: $\alpha$
	\item The upper limit to clip input tensor: $\beta$
	\item The zero point to be mapped to 0: $z$
	\item The scaling factor: $S$
\end{itemize}

To fully exploit $\mathrm{N}$-bit integer expression ability, $z$ and $S$ can be determined:
\begin{equation}
	\small
	\label{eq:S come from alpha, beta}
	z=\cfrac{\alpha+\beta}{2}
\end{equation}

\begin{equation}
	\small
	\label{eq:S come from alpha, beta}
	S=\cfrac{\beta-\alpha}{2^\mathrm{N}-2}
\end{equation}

The quantization function $Q$ uses a division to do linear mapping:
\begin{equation}
	\small
	\label{eq:quantization uniform}
	Q(\mathrm{act})=\mathrm{Int}(\frac{\mathrm{act}-z}{S})
\end{equation}

This uniform quantization function is simple and effective, but there are two problems. The first problem exists in both simulated quantization and integer-only quantization. The second problem only exists in integer-only quantization.
\subsection{Quantization Problem 1: Clipping Range}
\label{subsec:quantization problem: activation ranges}

In uniform quantization, the first problem is the determination of $\alpha$ an $\beta$: a unified clipping range is needed for all input activations, but the actual ranges vary among different batches of activations. Whether the quantization scheme is integer-only or not integer-only, this problem exists. If $[\alpha,\beta]$ covers a big range, the expressive ability of quantized tensor will be wasted; if $[\alpha, \beta]$ covers a small range, some float value will be clipped, causing performance degradation.

One method to gain $\alpha$ and $\beta$ is to set the minimum and maximum value of each activation to be $\alpha$ and $\beta$, and to sample their moving average in the quantization inference\cite{google}. 

The drawback of this method is that distribution of activation has long "tails"\cite{survey}, making $\alpha$ too "negative" and $\beta$ too "positive". The distribution of numbers near $\alpha$ and $\beta$ may be too sparse.

Another method is to use the percentile method\cite{percentile}, for example, use the range [$\alpha$, $\beta$] to cover only 95\% of the activation range. By stripping the long "tails", this method performs better.
The percentile method works well for 8 bit quantization. But in the training process, the calculation to gain $\alpha$ and $\beta$ may require sorting of tensor elements, which can cost a lot.

\subsection{Quantization Problem 2: Division}
\label{subsec:quantization problem: division}

The second problem in uniform quantization is the division operation in \eref{eq:quantization uniform}. Division is not a hardware-friendly operation. Even if the dividend and divisor are both integer, the calculation still require a lot of time and power. In the process of linear mapping, division is essential, causing most uniform quantization methods to be not integer-only.

\cite{google} proposed a method to approximate division. In this method, the division is replicated by multiplying a big integer and then doing right shifting:
\begin{equation}
	\small
	\label{eq:approx of s}
	1/S\approx M_S \times 2^{-m_S}
\end{equation}
So \eref{eq:quantization uniform} can be rewritten as:
\begin{equation}
	\small
	\label{eq:approx of s}
	Q(\mathrm{act})=\mathrm{Int}(({\mathrm{act}-z)}*M_S)\gg m_S)
\end{equation}

This method avoids division, but there's one drawback:
It contains big integer multiplication(32bit$\times$32bit), which still can cost a lot.

\subsection{Important empirical fact about Batch Normalization}
\label{subsec:important empirical fact about batch normalization}
BN layer is widely used in neural networks. By using an empirical fact of batch normalization layer, there is a subtle way to solve both problems illustrated in \ssref{subsec:quantization problem: activation ranges} and \ssref{subsec:quantization problem: division}. 

The purpose of BN layer is to force activations to follow standard Gaussian distribution.
Batch normalizing an activition $\mathrm{act}$ will be like:
\begin{equation}
	\small
	\label{eq:bn function}
	\mathrm{BN(act)=\cfrac{act-\mu(act)}{\sigma(act)}}
\end{equation}
where $\mu(.)$ is the mean function and $\sigma(.)$ is the standard deviation function. BN layer assumes that all activations follow Gaussian distribution. Subtracting mean and dividing standard deviation will force them to follow standard Gaussian distribution.

\begin{figure}[!t]
	\centering
	\includegraphics[width=0.5\textwidth]{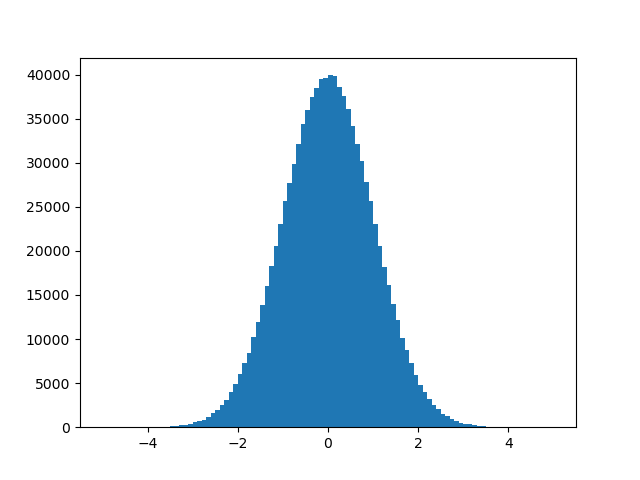}
	\caption{The histogram of 1 million float numbers following standard Gaussian distribution. Almost all numbers can be covered by range [-4, 4].
	}
	\label{fig:uniform}
\end{figure}

Here's the key empirical fact: one tensor following standard Gaussian distribution can be clipped by range $[-4, +4]$. \fref{fig:uniform} shows the histogram of 1 million float numbers following standard Gaussian distribution, only very few numbers fall out of the range. This empirical estimation can be directly used in quantization. If we want to quantize a tensor that follows standard Gaussian distribution, we can directly take $\alpha$ to be $-4$ and $\beta$ to be $+4$, without any calculation. It's also important that the absolute values of both $\alpha$ and $\beta$ are powers of 2, which will convert division operation to bit shifting operation. So, this empirical estimation kills two birds.

However, BN algorithm cannot be applied in integer-only network, because it contains division operation, which is not hardware-friendly. The method to address this is called Shift-based Batch Normalization(SBN).

\subsection{Shift-Based Batch Normalization\cite{bnn}.}
\label{subsec:shift-based batch normalization}
 To avoid float operation, SBN uses bit shifting to replicate division. 
In SBN layer, function $\mathrm{AP2}$ is used for this approximation:
\begin{equation}
	\small
	\label{eq:ap2 function}
	\mathrm{AP}(x)= \lceil \mathrm{log_2}(x)\rfloor
\end{equation}

After applying the $\mathrm{AP2}$ function, dividing $x$ can be approximated to right shifting $\mathrm{AP2}(x)$. This approximation is much rougher than that mentioned in \ssref{subsec:quantization problem: division}. The impact of this approximation will be illustrated later.

With this approximation, the SBN function is:
\begin{equation}
	\small
	\label{eq:sbn function}
	\mathrm{SBN}(\mathrm{act}_\mathrm{int})=(\mathrm{act_{int}}-\lceil \mu(\mathrm{act_{int}})\rfloor) \gg \mathrm{AP2(\sigma(act))}
\end{equation}

During training, the process should be simulated with float point, therefore floor operator will be used, the actual calculation process is like:
\begin{equation}
	\small
	\label{eq:sbn function float}
	\mathrm{SBN}(\mathrm{act_{int}})=\Bigg\lfloor\cfrac{\mathrm{act_{int}}-\lceil \mu(\mathrm{act_{int}})\rfloor}{\mathrm{ pow(2,AP2(\sigma(act_{int})))}}\Bigg\rfloor
\end{equation}

\begin{figure}[!t]
	\centering
	\includegraphics[width=0.5\textwidth]{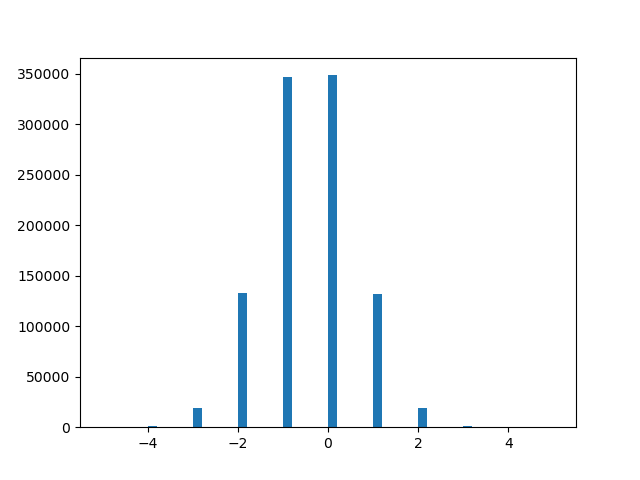}
	\caption{The histogram of the SBN result of 1 million float numbers following Gaussian distribution $(mean=100,std=1000)$.
 The result tensor is automatically quantized by floor operation.	}
	\label{fig:sbn}
\end{figure}

The SBN layer includes floor operation, so the output tensor is automatically quantized. \fref{fig:sbn} shows the SBN result of 1 million numbers that follows Gaussian distribution ($mean=100$, $std=1000$). SBN layer is directly used to replicate normal BN layer in order to reduce training cost in \cite{bnn}. In this extreme quantization network, SBN didn't bring any performance loss compared to normal BN.

\begin{figure}[!t]
	\centering
	\includegraphics[width=0.5\textwidth]{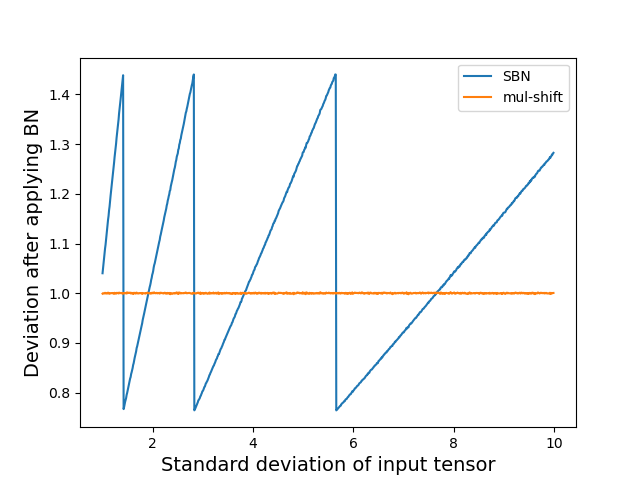}
	\caption{The standard of deviation after applying two methods(mul-shifting and merely shifting). On points landing on power of 2(i.e. $2^0, 2^1, 2^2...$), there's no approximation error. On points landing between power of 2(i.e. $2^{1.5},2^{2.5},2^{3.5}...$), the approximation error reaches maximum.}
	\label{fig:sbnapproximationerror}
\end{figure}

In fact, the SBN function will bring significant approximation error for replicating division by shifting. The approximation performance of SBN is far worse than the multiplying-shifting method mentioned in \ssref{subsec:quantization problem: division}.
\fref{fig:sbnapproximationerror} shows the approximation performance of two methods:
\begin{itemize}
	\item Replicating division with multiplying and shifting.
	\item Replicating division with merely shifting(SBN).
\end{itemize}

The ideal output standard deviation is 1. The results show that the first method almost has no approximation error and the deviation of output tensor is always nearly 1. But for SBN, the approximation error can be big. If the standard deviation of input tensor is near to power of 2(i.e. $2^1$, $2^2$, $2^3$...), there's no approximation error. But if the standard deviation lands between the powers of 2(i.e. $2^{1.5}$, $2^{2.5}$, $2^{3.5}$...), the output deviation will be $\sqrt{2}$ or $1/{\sqrt{2}}$. However, even with such big approximation error, the quantization scheme in \cite{bnn} works well.

\subsection{Combining SBN with Quantization}
\label{subsec:combining sbn with quantization}
\ssref{subsec:important empirical fact about batch normalization} shows that, the clipping range of one batch normalized tensor is easy to determine. Also, the range is easy to be quantized using shift operation. Therefore, we combine the hardware-friendly version of BN(which means SBN) with quantization, creating a new layer: Shift-based Batch Normalization Quantization(SBNQ).

In SBN, there is a shift operation. And in the quantization process, there is also a shift operation. These two shift operations can be composed into one. In order to achieve this fusion, we use a non-quantization version of SBN, which doesn't contain the floor operation.
\begin{equation}
	\small
	\label{eq:sbn function float}
	\mathrm{SBN}(\mathrm{act_{int}})=\cfrac{\mathrm{act_{int}}-\lceil \mu(\mathrm{act_{int}})\rfloor}{\mathrm{ pow(2,AP2(\sigma(act)))}}
\end{equation}

For the tensor already applied SBN, we still take the following parameter:

\begin{equation}
	\small
	\label{eq:s for tensor following std gaussian}
	\alpha=-4,	\beta=+4
\end{equation}
then $z$ and $S$ can be determined:
\begin{equation}
	\small
	\label{eq:s for tensor following std gaussian}
	z=\cfrac{\alpha+\beta}{2}=\cfrac{(-4)+(+4)}{2}=0
\end{equation}
\begin{equation}
	\small
	\label{eq:s for tensor following std gaussian}
	S=\cfrac{\beta-\alpha}{2^\mathrm{N}-2}=\cfrac{(-4)+(+4)}{2^{\mathrm{N}}-2}\approx \mathrm{pow(2,3-N)}
\end{equation}
therefore, the quantization SBNQ is:
\begin{equation}
	\small
	\label{eq:s for tensor following std gaussian}
	\begin{aligned}
		&\mathrm{SBNQ(act_{int})}\\\\
		=&\mathrm{Int}(\cfrac{\mathrm{SBN(act_{int})}-z}{S})\\\\
		=&\mathrm{Int}(\cfrac{\mathrm{\cfrac{\mathrm{act_{int}}-\lceil \mu(\mathrm{act_{int}})\rfloor}{\mathrm{ pow(2,AP2(\sigma(act_{int})))}}}-0}{\mathrm{pow(2,3-N)}})\\\\
		=& \mathrm{Int}(\cfrac{\mathrm{act_{int}}-\lceil \mu(\mathrm{act_{int}})\rfloor}{\mathrm{pow(2,N-3+AP2(\sigma(act_{int})))}})\\\\
		\approx& (\mathrm{act_{int}}-\lceil \mu(\mathrm{act_{int}})\rfloor)\gg (\mathrm{N}-3+\mathrm{AP2(\sigma(act_{int}))})
	\end{aligned}
\end{equation}

In the training process, quantization is simulated using float point numbers, and we store a moving average of $\mu$ and $\sigma$. In order to convert the network into an integer-only network, we create two persistent buffers in SBNQ layer:
\begin{equation}
	\small
	\label{eq:bias shift}
	\begin{aligned}
		\mathrm{bias}=&\lceil \mathrm{MovingAvg}(\mu(\mathrm{act_{int}}))\rfloor\\
		\mathrm{shift}=&N-3+\mathrm{AP2(MovingAvg(\sigma(act_{int})))}
	\end{aligned}
\end{equation}

So in the inference stage, the SBNQ function is very simple:
\begin{equation}
	\small
	\label{eq:s for tensor following std gaussian}
	\mathrm{SBNQ(act_{int})}=(\mathrm{act_{int}-bias})\gg \mathrm{shift}
\end{equation}
in which $\mathrm{bias}$ and $\mathrm{shift}$ are both integer.

\subsection{Quantization for Unsigned Activations and Weights}
\label{subsec:weight quantization}
For activations after $\mathrm{ReLU}$ layer, the SBNQ method cannot be applied because it doesn't follow Gaussian distribution. But we can assume that, before $\mathrm{ReLU}$ layer, the activation follows Gaussian distribution, then the $\mathrm{ReLU}$ layer can be fused into SBNQ layer. We can sample the value of $\mu$ and $\sigma$ before applying $\mathrm{ReLU}$ layer, and do quantization after $\mathrm{ReLU}$ layer.  And for unsigned quantization, the range of integers is $[0, 2^\mathrm{N}-1$], the value of $\alpha$ is $0$ instead of $-4$. Therefore, the scale factor will be:
\begin{equation}
	\small
	\label{eq:s for tensor following std gaussian}
	S=\cfrac{\beta-\alpha}{2^\mathrm{N}-1}=\cfrac{(+4)-0}{2^{\mathrm{N}}-1}\approx \mathrm{pow(2,2-N)}
\end{equation}

Therefore, the value of $\mathrm{shift}$ is different from \eref{eq:bias shift}:
\begin{equation}
	\small
	\label{eq:s for tensor following std gaussian}
	\begin{aligned}
		\mathrm{shift}&=N-2+\mathrm{AP2(moving\_average(\sigma(act_{int})))}\\
	\end{aligned}
\end{equation}

And the $\mathrm{ReLU}$ layer is fused into SBNQ before shifting:
\begin{equation}
	\small
	\label{eq:s for tensor following std gaussian}
	\begin{aligned}	
		\mathrm{SBNQ(act_{int})}&=\mathrm{ReLU}(\mathrm{act_{int}-bias})\gg \mathrm{shift}
	\end{aligned}
\end{equation}

The technique of SBNQ can also be used in the quantization of weights. We assume that all weight follows Gaussian distribution. If the weight needs be quantized into $\mathrm{N}$-bit integer, we will initialize the weight $w$ with random numbers with specified standard deviation value:

\begin{equation}
	\small
	\label{eq:s for tensor following std gaussian}
	\sigma(w_\mathrm{init})=\mathrm{pow}(2,\mathrm{N}-3)
\end{equation}
then we can use range $\mathrm{[-2^{N-1}+1,2^{N-1}-1]}$ to accurately clip the weight. To quantize $w$, only rounding will be needed:
\begin{equation}
	\small
	\label{eq:s for tensor following std gaussian}
	\begin{aligned}
		Q(w)&=\lceil w\rfloor
	\end{aligned}
\end{equation}



\section{Experiments}
\label{sec:experiments}
We applied this integer-only quantization scheme to train some classical datasets. Because we have limited computing resources (one NVIDIA V100), some results are not as good as SOTA results. Therefore, we carried out controlled experiments on floating networks and integer-only networks. In one controlled experiment, one float network and one quantized network shared the same topological architecture. The quantization precision is 4-bit\footnote{The results for 8 bit quantization is worse than 4 bit quantization. It's weird, but we haven't solve this problem yet.}. The performance is shown in \tref{tab:experiments}.

The results for MNIST, CIFAR10 and CIFAR100 show that the quantization network even performs better. Maybe during the STE back propagation, the quantization noise becomes one kind of regularizer, which helps to avoid over-fitting.

For ImageNet, we constructed a simplified VGG net with fewer fully connected layers. The result shows that, on tough tasks, SBNQ will have observable performance degradation(about 12\%). But the loss may be tolerated, because 4 bit quantization is extreme and may bring great inference efficiency.

\begin{table}
	\centering
	\normalsize
\begin{tabular}{|c|l|l|l|l|}
	\hline
	\cellcolor[HTML]{FFFFFF}                          & \multicolumn{2}{c|}{{\color[HTML]{FD6864} float}} & \multicolumn{2}{c|}{{\color[HTML]{00D2CB} integer-only}} \\ \cline{2-5} 
	\multirow{-2}{*}{\cellcolor[HTML]{FFFFFF}dataset} & top-1                   & top-5                   & top-1                       & top-5                      \\ \hline
	MNIST                                             & 99.35\%                 & 99.99\%                 & 99.39\%                     & 99.99\%                    \\ \hline
	CIFAR10                                           & 86.45\%                 & 98.65\%                 & 86.92\%                     & 98.47\%                    \\ \hline
	CIFAR100                                          & 55.70\%                 & 77.69\%                 & 55.71\%                     & 75.76\%                    \\ \hline
	\multicolumn{1}{|l|}{IMAGENET}                    & 54.89\%                 & 78.15\%                 & 42.77\%                     & 71.28\%                    \\ \hline
\end{tabular}
	\vspace*{8pt plus 3pt minus 2pt}
	\caption{The experiment results of float network and integer-only network(4bit).}
	\label{tab:experiments}
\end{table}


\section{Discussion}
\label{sec:discussion}

We proposed a quantization scheme that can avoid big integer multiplication, which can be used in low precision neural network training and inference. In the whole integer-only inference process, only these integer operations are used:
\begin{itemize}
	\item Multiplication of 4 bit integer, producing 8 bit integer.
	\item Addition of 8 bit and 32 bit integer, producing 32 bit integer.
	\item Right shifting of 32 bit integer, producing 4 bit integer.
\end{itemize}

We tried 4-bit quantization on some simple datasets(MNIST, CIFAR10, CIFAR100), and there was no quantization loss. We also tried 4-bit quantization on big dataset(ImageNet). The experiment result shows that there is about 12\% performance loss, but due to the inference efficiency of 4-bit quantization, this loss may be tolerable in some application scenarios.

\section{Future work}
\label{sec:conclusions}

 We are trying to design a RISC-V co-processor using hardware-architecture co-designing. The co-processor will have least hardware operations and bring most inference efficiency.

{
\bibliographystyle{plain}
\bibliography{ref.bib}
}

\end{document}